\documentclass[10pt,conference]{IEEEtran}
\IEEEoverridecommandlockouts 
\usepackage{amsmath,amssymb}
\usepackage{graphicx}
\usepackage{cite}
\usepackage{url}
\usepackage{xcolor}
\usepackage[colorlinks=true,urlcolor=blue]{hyperref}
\usepackage{booktabs}
\usepackage{multirow}
\usepackage{siunitx}
\usepackage{booktabs}
\usepackage{tabularx}
\usepackage{adjustbox}
\usepackage{float}
\usepackage{makecell}
\usepackage{comment}

\definecolor{mypink}{RGB}{230,0,120} 
\title{MaskAttn-SDXL: Controllable Region-Level Text-to-Image Generation}

\author{\IEEEauthorblockN{
Yu Chang\textsuperscript{1,*},
Jiahao Chen\textsuperscript{2,*},
Anzhe Cheng\textsuperscript{1,*},
Paul Bogdan\textsuperscript{1,\textdagger}}

\IEEEauthorblockA{\textsuperscript{1}University of Southern California, Los Angeles, CA, USA}
\IEEEauthorblockA{\textsuperscript{2}University of Toronto, Toronto, Canada}

\thanks{\textsuperscript{*}Equal contribution.}
\thanks{\textsuperscript{\textdagger} Corresponding author: Paul Bogdan (email: \href{mailto: pbogdan@usc.edu}{pbogdan@usc.edu}).}}

\begin{document}

\maketitle

\begin{abstract}
Diffusion models have achieved strong results in text-to-image generation, but important limitations remain as prompts become more structured and multi-object. On the architecture side, U-Net backbones are efficient and stable, yet their locality makes global coordination harder, while Transformer-based diffusion models improve global interactions but at substantially higher compute and memory cost.
In parallel, compositional reliability remains weak: models often mix attributes across objects, violate spatial relations, or omit requested entities, and these errors are not reliably reflected by global metrics such as FID or CLIP-based scores.
To address these issues without changing the SDXL pipeline, we propose MaskAttn-SDXL, a plug-in module that injects token-conditioned spatial gating into cross-attention logits before softmax. The gating sparsifies token-to-location interactions to suppress irrelevant bindings while preserving the pretrained backbone and standard sampling process, requiring no external supervision or inference-time editing. 
\smallskip
\\
\noindent\textcolor{blue}{\textbf{Code:} \url{https://github.com/ychang64/MaskAttn-SDXL}}

\end{abstract}

\begin{IEEEkeywords}
mask attention, diffusion models, generative modeling, spatial control
\end{IEEEkeywords}


\section{Introduction}
\label{sec:intro}

Diffusion models have emerged as a cornerstone of generative modeling, advancing applications across modalities including text-to-image synthesis, inpainting, audio generation, and video prediction~\cite{podell2024sdxl}. Their iterative denoising paradigm enables high-fidelity outputs with fine-grained control, and recent large-scale variants such as SDXL have demonstrated strong text alignment and visual realism~\cite{ho2020denoising,rombach2022ldm}. Yet key structural and semantic challenges remain, especially as prompts grow more complex.

A central limitation lies in the architectural foundation of diffusion backbones. The dominant class, convolutional U-Nets, benefit from strong spatial locality and inductive bias, enabling efficient and stable training~\cite{ronneberger2015unet,podell2024sdxl}. However, limited receptive fields hinder modeling of long-range dependencies and global composition~\cite{luo2016understanding,wang2018non}. Transformer-based diffusion models such as DiT~\cite{peebles2023scalable} and JiT~\cite{li2025back} address this with global self-attention, enabling richer token interactions across spatial regions~\cite{peebles2023scalable,vaswani2017attention}. Nevertheless, they require more compute, incur higher memory costs, and often underperform at capturing fine-grained local structure~\cite{dosovitskiy2020image}. As a result, a persistent trade-off remains between efficiency and global coherence.

Beyond backbone design, compositional reliability in text-to-image generation remains a critical challenge~\cite{huang2023t2i,chefer2023attend,ghosh2023geneval}. With prompts containing multiple entities, attributes, and spatial relations, even state-of-the-art models often fail to maintain token-level consistency. Common failure modes include attribute leakage, where colors or properties are misassigned~\cite{huang2023t2i,zarei2024understanding}; spatial violations such as inverted left--right positioning~\cite{huang2023t2i,ghosh2023geneval}; and object omissions~\cite{ghosh2023geneval,chefer2023attend}. These errors persist even in large-scale models trained on diverse corpora and are rarely captured by global metrics such as FID or CLIP~\cite{heusel2017gans,radford2021learning}. At the core is cross-token interference, where textual concepts compete for shared spatial attention, producing entangled or mismatched visual outputs.

\begin{figure}[!t]
  \centering
  \includegraphics[width=\columnwidth]{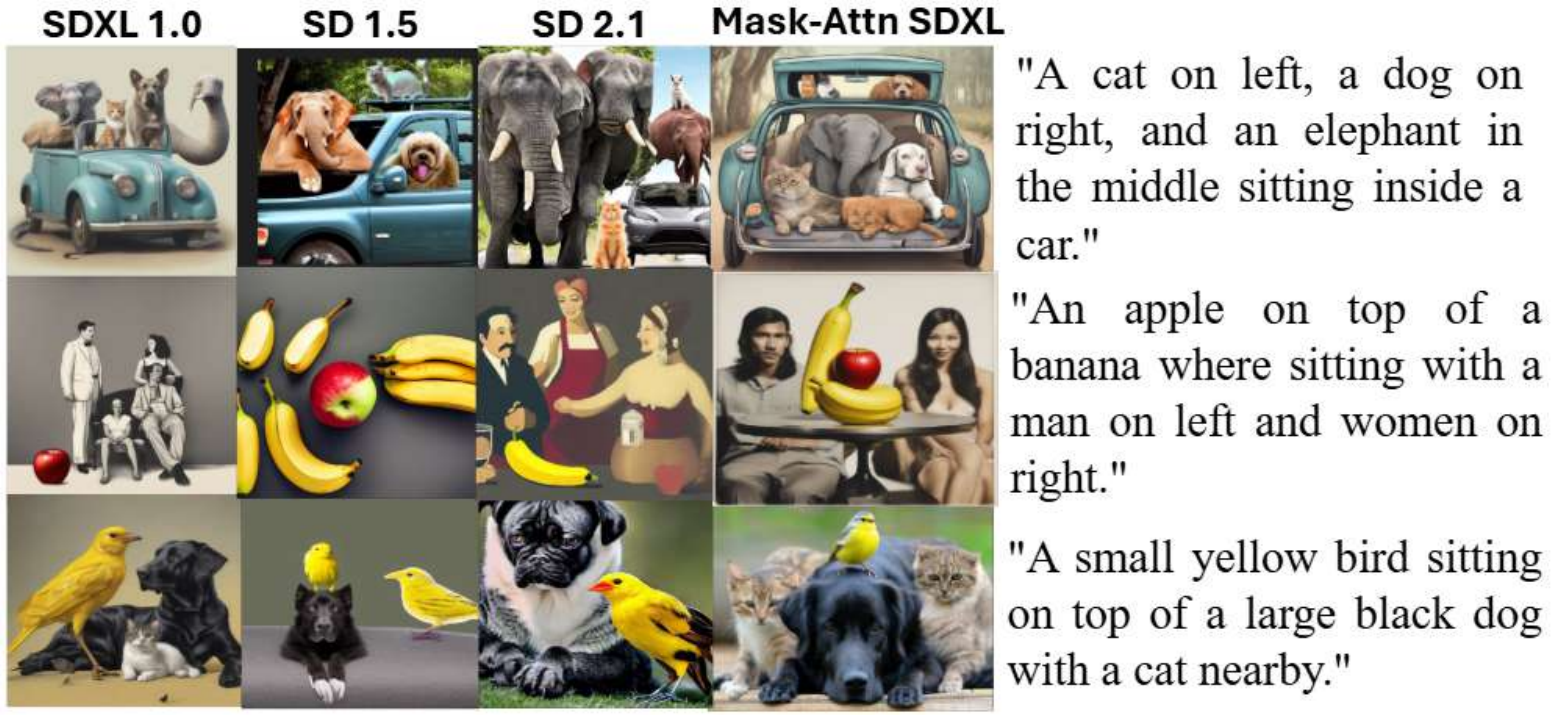}
  \caption{Qualitative comparison showing that, on multi-object spatial prompts, MaskAttn-SDXL reduces object overlap and improves compositional correctness relative to SDXL, SD-1.5, and SD-2.1.}
  \label{fig:teaser}
\end{figure}

\begin{figure*}[!t]
  \centering
  \includegraphics[width=\textwidth]{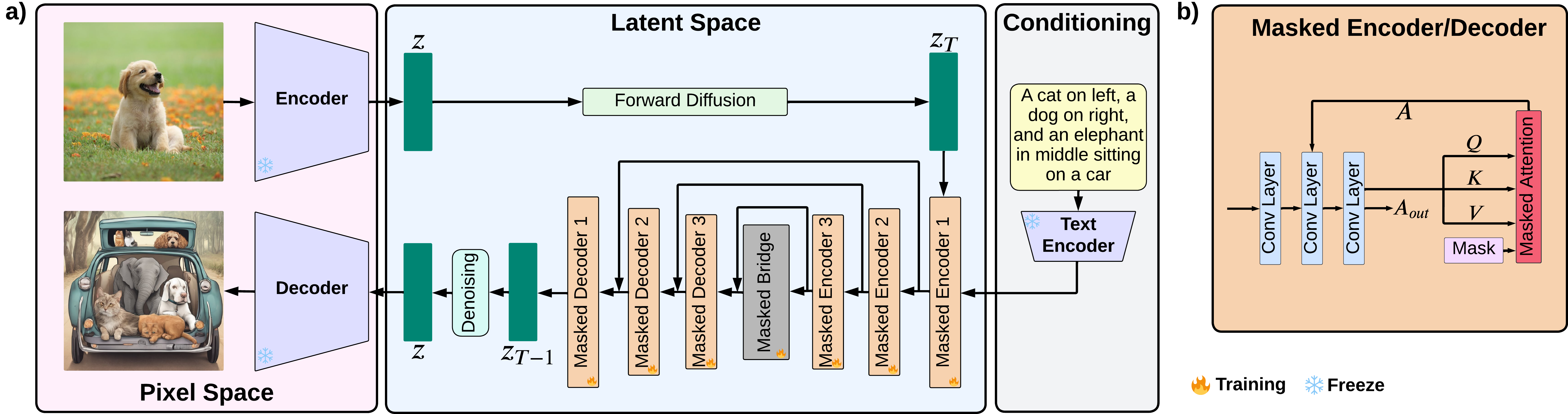}
  \caption{\textbf{Overview of MaskAttn-SDXL.}(a)Images are encoded to latent $z$, noised to $z_T$ by forward diffusion, then denoised by MaskAttn Unet, text encoder provides conditioning. (b)The mechanism of Masked Encoder/Decoder. The attention is masked and combined with $A_{out}$ to update latent features.}
  \label{fig:flowchart}
\end{figure*}

To address these drawbacks, we propose \textbf{MaskAttn-SDXL}, a lightweight enhancement that injects learnable, token-conditioned masked cross-attention into the logits space of Stable Diffusion XL (SDXL). At selected layers of the SDXL U-Net, a binary spatial mask modulates token-to-location attention, suppressing spurious interactions and sharpening token--region alignments. Unlike prior approaches~\cite{bar2023multidiffusion,li2023gligen,zhang2023adding,cheng2025maskattn,bogdan2026ai}, our method requires no auxiliary inputs such as bounding boxes, adds negligible parameters, and leaves the pretrained backbone and inference pipeline unchanged. This improves compositional grounding in a text-only setting without sacrificing efficiency or deployment flexibility.

Our contributions are summarized as follows:
\begin{itemize}
    \item We propose a logit-space masked attention mechanism that sparsifies token-to-location connections in SDXL, improving compositional alignment.
    \item Our method improves spatial and attribute fidelity under complex prompts, with stronger performance on compositional benchmarks such as T2I-CompBench++ and GenEval.
    \item MaskAttn-SDXL achieves these gains with minimal runtime and memory overhead while remaining compatible with pretrained SDXL and standard text-only workflows.
\end{itemize}

\section{Related Work}

\noindent\textbf{Variety Backbones.}
Most text-to-image diffusion models use CNN-based U-Nets as the denoising backbone, as in DDPM-style and latent diffusion systems~\cite{ho2020ddpm,ronneberger2015unet,rombach2022ldm}.
This design remains dominant in high-resolution pipelines such as SDXL~\cite{podell2024sdxl} because convolutional locality provides a strong inductive bias for stable training and efficient scaling.
Recent work instead explores more globally attentive backbones built on self-attention for long-range dependency modeling, including diffusion transformers (DiT)~\cite{peebles2023dit} and Transformer-based text-to-image generators such as PixArt-$\alpha$~\cite{chen2023pixart}.
These architectures can improve global coordination, but often require more compute and sacrifice convolutional priors beneficial for fine local structure.

\noindent\textbf{Composition-oriented interventions.}
A complementary line of work improves compositional behavior without retraining by intervening at inference time, often through direct cross-attention manipulation to reduce cross-token competition.
Prompt-to-Prompt~\cite{hertz2022prompt} exploits the link between cross-attention maps and spatial layout, enabling localized edits by reusing or replacing token-wise attention correspondences across denoising steps.
Attend-and-Excite~\cite{chefer2023attend} addresses failures such as missing objects by inspecting attention mass for specified tokens and applying sampling-time updates to increase token coverage, improving object inclusion at the cost of extra refinement.
Structured Diffusion Guidance~\cite{feng2023structured} incorporates linguistic structure into the guidance signal to better enforce multi-entity constraints while keeping pretrained weights fixed.
Overall, these training-free approaches preserve the original backbone and are easy to apply, but they can add inference-time overhead and may vary across prompts and layouts because they rely on sampling-time heuristics.

\section{Method}
\label{sec:pagestyle}
\subsection{Architecture Overview}

As shown in Fig.~\ref{fig:flowchart}a, MaskAttn-SDXL extends the SDXL latent diffusion pipeline with masked cross-attention gating. MaskAttn-SDXL follows the standard SDXL latent-diffusion pipeline and keeps the overall generation procedure unchanged. Our modification is applied only inside the SDXL U-Net cross-attention blocks. By default, we insert it at selected mid-resolution blocks, which provide a good balance between semantic abstraction and spatial precision. At each selected site, lightweight masking heads predict a token-conditioned spatial gate from the current latent feature map and token embedding. This gate is injected into the cross-attention logits before softmax, directly modulating token-to-location competition (Fig.~\ref{fig:flowchart}b). Intuitively, the mask suppresses semantically irrelevant token--region interactions, reducing cross-token interference in prompts with multiple entities and relations.

Importantly, MaskAttn-SDXL is designed as a plug-in extension: all pretrained SDXL components (U-Net backbone weights, text encoders, VAE, and the sampling procedure) remain frozen, and only the masking heads are trained. As a result, the method preserves SDXL’s inference workflow while adding a small, targeted module to improve token–region alignment under text-only prompting.

\subsection{Mask Attention Gating}

The core of our method is the masked attention module. 
To construct the mask matrix $M$, we attach a lightweight gate head, $f(\cdot)$, at each cross-attention site. The gate head is a small spatial predictor (e.g., a shallow conv/MLP head) operating at the native resolution of $X$; it is shared across diffusion timesteps and adds only a small parameter footprint relative to the frozen SDXL backbone.
The gate head $f$ takes the current feature map $X \in \mathbb{R}^{H \times W \times C}$ and the $t$-th token embedding $e_t$ as input, and outputs a token-conditioned spatial probability map $\hat G_{t} \in (0,1)^{H \times W}$:

\begin{equation}
\hat G_{t} = \sigma\!\big(f(X, e_t)\big),
\label{eq:gate_prob}
\end{equation}

where $\sigma(\cdot)$ is the sigmoid activation function. Intuitively, the probability map $\hat{G}_{t}(x,y)$ represents the model's belief in how strongly token $t$ should contribute to the image generation at spatial coordinate $(x, y)$ in layer $l$.
The learned maps $\{G_{t}\}$ are directly visualizable as token-wise spatial masks.

We then binarize this continuous probability map into a hard gate $G_t$ using a threshold of 0.5, employing a straight-through estimator (STE) to allow gradient flow during training. The threshold of 0.5 is used as the default decision boundary for sigmoid outputs, yielding a hard spatial gate that enables explicit token-location suppression.

This binary gate is subsequently converted into the additive mask matrix $M$. For each spatial location $i$ and token $t$:
\begin{equation}
G_{t}(i) =
\begin{cases}
1, & \hat{G}_{t}(i) > 0.5, \\
0, & \text{otherwise},
\end{cases}
\label{eq:gate_bin}
\end{equation}

and then convert it into the additive mask matrix $M$:
\begin{equation}
M(i,t) =
\begin{cases}
0, & G_{t}(i) = 1, \\
-\infty, & G_{t}(i) = 0.
\end{cases}
\label{eq:mask_matrix}
\end{equation}
In implementation, we approximate $-\infty$ with a large negative constant for numerical stability. This bias is directly added to the cross-attention logits below, effectively suppressing tokens at locations where the gate is off. And the output of the attention operation for a given head is then combined across all heads (as in multi-head cross--attention) and added to the original input via a residual connection.

Given queries $Q \in \mathbb{R}^{N \times d}$, keys $K \in \mathbb{R}^{T \times d}$, and values $V \in \mathbb{R}^{T \times d}$ at a given specific U-Net layer (where $N$ is the number of spatial locations and $T$ is the number of text tokens), we integrated the learnable, additive mask matrix $M \in \mathbb{R}^{N \times T}$ directly to the attention logits. The masked attention is computed as:

\begin{equation}
\text{MaskAttn}(Q, K, V; M) 
= \text{Softmax}\!\left(\frac{Q K^{\top}}{\sqrt{d}} + M\right) V
\label{eq:attn_bias}
\end{equation}

The matrix $M \in \mathbb{R}^{N \times T}$ provides an entry-wise gate over token--location interactions: each element $M(i,t)$ biases the attention logit for spatial location $i$ attending to token $t$ before the softmax normalization.
Since $M$ is added in logit space, setting $M(i,t)=-\infty$ (Eq.~\ref{eq:mask_matrix}) forces the corresponding softmax weight to be zero, i.e., the model assigns zero attention probability to token $t$ at location $i$. This yields hard token-to-location suppression without changing the standard attention operator.

Let $A$ represent the output of the masked multi-head attention 
(after head aggregation). This output is passed through a two-layer 
feed-forward network (FFN) with a GELU activation, and combined with 
the residual connection:

\begin{equation}
A_{\text{out}} = \text{GELU}(AW_{1} + b_{1})W_{2} + b_{2} + A
\end{equation}

where $W_{1}, W_{2}$ are the weight matrices and $b_{1}, b_{2}$ are the 
corresponding biases of the FFN. The resulting $A_{\text{out}}$ constitutes 
the final output of the cross--attention block. This design allows the residual 
FFN to refine the masked attention features by incorporating global context, 
while maintaining the original information through the skip connection~\cite{vaswani2017attention}.

\section{Experiments}
\label{sec:typestyle}



 To rigorously assess our approach and enable meaningful, controlled comparisons with state-of-the-art diffusion models, we evaluate MaskAttn-SDXL along two complementary axes:
(i) \emph{quality evaluation} on MS-COCO 2014 and Flickr30k to measure global fidelity and text--image alignment; and
(ii) \emph{compositional benchmarks} that explicitly test spatial/attribute binding under structured multi-object prompts, including T2I-CompBench++ (Spatial; UniDet)~\cite{huang2025t2icompbenchpp} and GenEval~\cite{ghosh2023geneval}.
Together, these evaluations quantify both overall generation quality and compositional correctness 
under text-only prompting.
We enforce uniform training and evaluation hyperparameters across methods unless a setting is model-specific by design, in which case we state it explicitly.

\begin{table}[H]
\centering
\caption{Quality (fidelity) results on MS-COCO and Flickr30k.}
\label{tab:quality_only}
\setlength{\tabcolsep}{5.2pt}
\renewcommand{\arraystretch}{1.10}
\resizebox{0.48\textwidth}{!}{%
\begin{tabular}{l cc cc}
\toprule
& \multicolumn{4}{c}{\textbf{Quality (Fidelity)}} \\
\cmidrule(lr){2-5}
\textbf{Method} 
& \multicolumn{2}{c}{\textbf{MS-COCO}} 
& \multicolumn{2}{c}{\textbf{Flickr30k}} \\
\cmidrule(lr){2-3}\cmidrule(lr){4-5}
& \textbf{FID}$\downarrow$ 
& \textbf{CLIP}$\times 10^{2}\uparrow$
& \textbf{FID}$\downarrow$ 
& \textbf{CLIP}$\times 10^{2}\uparrow$ \\
\midrule
SD-1.5                         & \textbf{24.01} & 30.17 & 203.80 & 32.31 \\
SD-2.1-base                    & 24.25 & 31.32 & \textbf{202.60} & 32.71 \\
PixArt-$\alpha$                & 25.68 & 31.49 & 210.30 & 32.12 \\
PixArt-$\Sigma$                & 25.23 & 31.55 & 209.12 & 32.01 \\
SDXL                           & 25.77 & 31.53 & 209.80 & 33.03 \\
\textbf{MaskAttn-SDXL (mid-res, default)} 
                               & 24.57 & \textbf{31.75} & 206.98 & \textbf{33.54} \\
\bottomrule
\end{tabular}%
}
\end{table}

\subsection{Experimental Setup}

MaskAttn-SDXL is fine-tuned on COCO captions while freezing the base SDXL diffusion weights. We use COCO \textit{train2014} and construct 200k image--caption pairs biased toward multi-entity captions (at least two noun phrases). The module is trained for 100k steps with an effective batch size of 16 at $512{\times}512$ using AdamW (learning rate $1{\times}10^{-4}$, weight decay $0.01$, $\beta_1{=}0.9$, $\beta_2{=}0.999$), 1k warmup steps with cosine decay, gradient clipping at 1.0, and mixed precision. To adapt to high-resolution of SDXL, we perform an additional 10k fine-tuning steps at $1024{\times}1024$ with batch size 8.


After training, we generate images using each method’s standard sampling pipeline.
For SD-family models (SD-1.5, SD-2.1-base, and SDXL), we match the sampling procedure (diffusion steps and noise schedule) and vary only model weights/architecture, evaluating each model at its native resolution ($512{\times}512$ for SD-1.5/SD-2.1-base; $1024{\times}1024$ for SDXL/MaskAttn-SDXL).
For PixArt-$\alpha$/$\Sigma$~\cite{chen2024pixartalpha,chen2024pixartsigma} and compositional/editing baselines (Composable Diffusion (SD-2.1)~\cite{liu2022composable}, Structured Diffusion (SD-2.1)~\cite{feng2023structured}, and Attend-and-Excite (SD-2.1)~\cite{chefer2023attend}), we follow method-recommended inference settings while keeping the sampling budget comparable.

\begin{table}[t]
\centering
\caption{Compositional benchmark results.}
\label{tab:composition_only}
\setlength{\tabcolsep}{1pt}
\begin{tabular}{lcc}
\toprule
\textbf{Method} 
& \thead{CompBench++ Spatial\\(UniDet)$\uparrow$}
& \thead{GenEval$\uparrow$} \\
\midrule
SD-1.5                         & 0.1352 & 0.49 \\
SD-2.1-base                    & 0.1423 & 0.56 \\
Composable SD (v2)            & 0.0750 & 0.53 \\
Structured SD (v2)              & 0.1333 & 0.55 \\
SDXL                           & 0.2090 & 0.62 \\
PIXART-$\alpha$                   & 0.2015 & 0.63\\
Attend-and-Excite SD (v2)             & 0.1471 & 0.54 \\
\textbf{MaskAttn-SDXL (mid-res, default)} 
                               & \textbf{0.2248} & \textbf{0.66} \\
\bottomrule
\end{tabular}%
\end{table}

\subsubsection{Datasets}
For quality evaluation, we use MS-COCO and Flickr30k. 
MS-COCO is our primary benchmark: we use 3{,}000 captions from the \texttt{val2014} split for generation and evaluation. 
We further evaluate generalization on Flickr30k using 500 sampled captions. 
Compositional performance is assessed on dedicated structured benchmarks (T2I-CompBench++ and GenEval), which provide multi-object prompts with explicit spatial/attribute constraints.

\subsubsection{Evaluation Metrics}
We report both global fidelity/alignment metrics and compositional benchmarks. The Fr\'echet Inception Distance (FID) quantifies perceptual quality by comparing generated and real image distributions in a deep feature space. Precision and Recall measure diversity and coverage of the generated distribution. The CLIP Score measures image--text semantic alignment.

However, global metrics alone can miss object-level compositional errors, such as violated spatial relations. Therefore, we additionally report compositional benchmarks that explicitly target spatial correctness, including T2I-CompBench++ (Spatial) and GenEval. These benchmarks better capture token-to-region faithfulness beyond overall realism, and can reveal failures that remain competitive under FID/CLIP despite violating the prompt structure.

\subsection{Baseline Models}

We compare MaskAttn-SDXL against a set of widely used text-to-image baselines. 
Our primary backbone comparisons focus on latent-diffusion models with cross-attention conditioning: SD-1.5, SD-2.1-base, and SDXL~\cite{rombach2022ldm,podell2024sdxl}. 
These models share the same latent-diffusion formulation and conditioning mechanism, while differing in training data and model scale, enabling a controlled test of whether the proposed gating improves compositional reliability beyond gains attributable to scaling or updated training corpora. 
To broaden coverage beyond the SD family, we additionally include strong non-SD, DiT-based baselines (PixArt-$\alpha$/$\Sigma$) and representative compositional/editing methods (Composable Diffusion , Structured Diffusion, and Attend-and-Excite) in the corresponding evaluations.

\begin{figure}[t]
  \centering
  \includegraphics[width=0.9\linewidth]{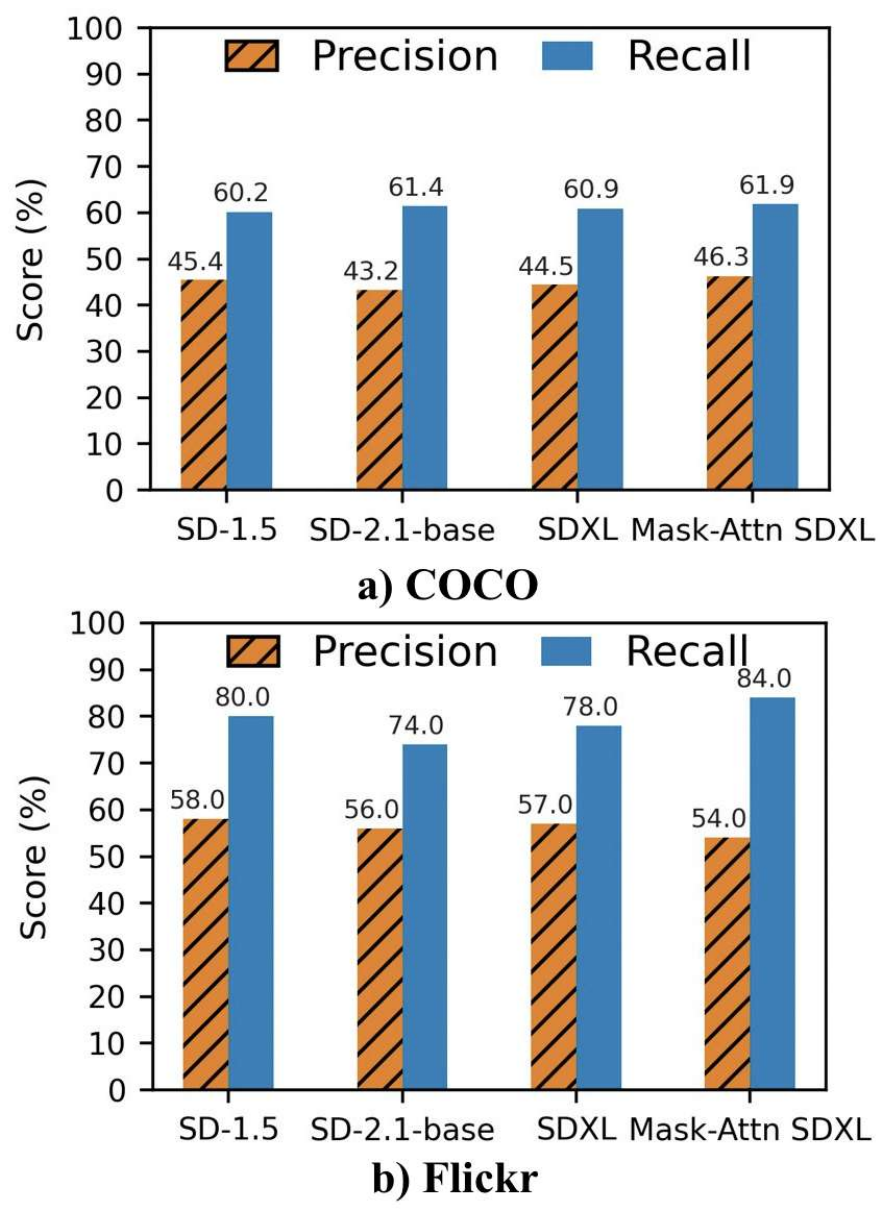}
  \caption{Precision and Recall comparison of different methods on (a) MS-COCO val2014 and (b) Flickr30k.}
  \label{fig:precision_recall}
\end{figure}

\begin{table}[H]
\centering
\caption{Computational overhead under fixed sampling settings.}
\label{tab:overhead}
\vspace{-4pt}
\setlength{\tabcolsep}{1.5pt}
\renewcommand{\arraystretch}{0.8}
\footnotesize
\begin{tabular}{lccc}
\toprule
\textbf{Method} & \textbf{Params (B)} & \textbf{Mem (GB)} & \textbf{Lat. (s/img)$\downarrow$} \\
\midrule
SDXL & 2.60 & 12.4 & 1.00 \\
SDXL + ControlNet (Canny) & 3.00 & 14.2 & 1.18 \\
SDXL + GLIGEN(box grounding) & 2.62 & 13.0 & 1.10 \\
\textbf{MaskAttn-SDXL} & 2.60 & 12.7 & 1.03 \\
\bottomrule
\end{tabular}
\vspace{-4pt}
\end{table}


\begin{table*}[t]
\small
\centering
\caption{Ablations on gating location and placement. Block A studies stage (resolution) selection; Block B studies module placement.}
\label{tab:ablations_where_to_gate}
\setlength{\tabcolsep}{6.0pt}
\renewcommand{\arraystretch}{1.08}
\begin{tabular}{lcccc}
\toprule
\textbf{Setting} 
& \textbf{CompBench++ Spatial (UniDet)}$\uparrow$
& \textbf{GenEval}$\uparrow$
& \textbf{FID}$\downarrow$
& \textbf{CLIP}$\times 10^{2}\uparrow$ \\
\midrule
\multicolumn{5}{l}{\textbf{Block A: Stage gating (resolution)} \hfill \textit{(3000 images)}}\\
\midrule
High-res blocks   & 0.2212 & 0.64 & 24.80 & 31.55 \\
\textbf{Mid-res blocks (default)} & \textbf{0.2248} & \textbf{0.66} & \textbf{24.57} & \textbf{31.75} \\
Low-res blocks    & 0.2235 & 0.65 & 24.68 & 31.62 \\
All blocks        & 0.2080 & 0.60 & 25.35 & 31.10 \\
\midrule
\multicolumn{5}{l}{\textbf{Block B: Placement (encoder / decoder)} \hfill \textit{(500 images)}}\\
\midrule
Encoder-only      & -- & -- & 109.65 & 30.54 \\
Decoder-only      & -- & -- & 109.83 & 30.41 \\
Full (encoder+decoder) & -- & -- & \textbf{109.05} & \textbf{31.02} \\
\bottomrule
\end{tabular}
\end{table*}

\subsection{Results}
Tab.~\ref{tab:quality_only} and Fig.~\ref{fig:precision_recall} report caption-based fidelity and alignment results on MS-COCO and Flickr30k (FID/CLIP and Precision--Recall). 
Tab.~\ref{tab:composition_only} reports compositional correctness on structured benchmark suites (T2I-CompBench++ Spatial and GenEval), which explicitly evaluate spatial/attribute binding beyond global image realism.

\textbf{Caption-based evaluation.}
On MS-COCO captions, MaskAttn-SDXL improves over SDXL in both fidelity and alignment.
As reported in Table~\ref{tab:quality_only}, MaskAttn-SDXL increases CLIP from 31.53 to 31.75 and reduces FID from 25.77 to 24.57.
Figure~\ref{fig:precision_recall}a further shows that MaskAttn-SDXL shifts the Precision--Recall profile relative to SDXL (Precision 45.4$\rightarrow$46.3, Recall 60.9$\rightarrow$61.9), consistent with reduced cross-token interference during generation.
Among the broader baselines in Table~\ref{tab:quality_only} (SD-1.5/SD-2.1-base and PixArt-$\alpha$/$\Sigma$), MaskAttn-SDXL achieves the strongest CLIP while improving FID relative to SDXL at $1024{\times}1024$.

We also observe that the $512{\times}512$ backbones (SD-1.5/SD-2.1-base) can achieve lower zero-shot FID on COCO captions, a phenomenon previously noted to not necessarily track human preference or compositional correctness for high-resolution models~\cite{podell2024sdxl}. In our setting, the more relevant comparison is against SDXL under matched sampling and resolution, where MaskAttn-SDXL consistently improves both fidelity and alignment.

The same trend holds on Flickr30k captions. MaskAttn-SDXL improves CLIP from 33.03 to 33.54 and reduces FID from 209.80 to 206.98 (Table~\ref{tab:quality_only}). Figure~\ref{fig:precision_recall}b shows a marked Recall increase for MaskAttn-SDXL (78.0$\rightarrow$84.0) with a moderate Precision decrease (57.0$\rightarrow$54.0), indicating improved coverage of valid modes while maintaining comparable fidelity.

\textbf{Compositional benchmark.}
Separately, we evaluate compositional correctness on structured benchmark suites with their own prompt sets. As reported in Table~\ref{tab:composition_only}, MaskAttn-SDXL achieves the best overall performance, improving over SDXL on T2I-CompBench++ Spatial (UniDet) from 0.2090 to 0.2248 and on GenEval from 0.62 to 0.66, and also outperforming SD-1.5, SD-2.1-base, PixArt-$\alpha$, and compositional/editing baselines including Composable Diffusion (SD-2.1), Structured Diffusion (SD-2.1), and Attend-and-Excite (SD-2.1).

\textbf{Efficiency and overhead.}
 Table~\ref{tab:overhead} quantifies computational overhead. MaskAttn-SDXL adds 3.6M parameters and incurs small memory and latency increases over SDXL, supporting the claim that the approach is lightweight yet effective, consistent with the improvements in Table~\ref{tab:composition_only}.

\textbf{Ablations.}
Table~\ref{tab:ablations_where_to_gate} isolates where masking is most effective. In Block A (stage selection), mid-resolution gating yields the best CompBench++ Spatial score (0.2248), supporting our default choice. In Block B (encoder/decoder placement), we report FID/CLIP as a compute-controlled diagnostic: these placement runs are measured on a smaller sample (500 images) and are primarily intended to attribute quality/alignment changes to the insertion location. Composition benchmarks (CompBench++/GenEval) are more evaluation-intensive; we therefore prioritize them for the full model and stage-selection ablations (3000 images).

\begin{figure}[t]
\centering
\setlength{\tabcolsep}{1.0pt}
\renewcommand{\arraystretch}{0.75}
\resizebox{0.92\linewidth}{!}{%
\begin{tabular}{@{}>{\raggedleft\arraybackslash}m{0.11\linewidth} *{3}{m{0.27\linewidth}}@{}}
\toprule
& \multicolumn{3}{c}{\shortstack[c]{\emph{`A red dragon on the left and}\\ \emph{a blue dragon on the right, cinematic shot'}}} \\
\midrule
\rotatebox[origin=c]{90}{\textbf{SD 1.5}} &
  \includegraphics[width=0.98\linewidth]{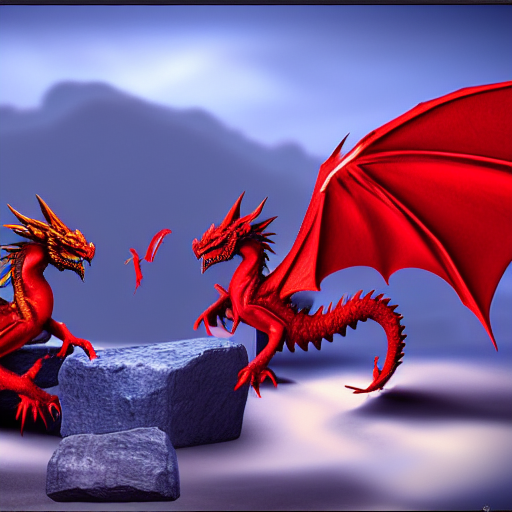} &
  \includegraphics[width=0.98\linewidth]{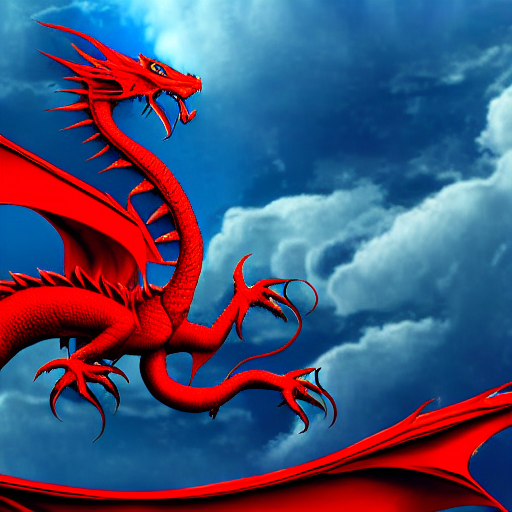} &
  \includegraphics[width=0.98\linewidth]{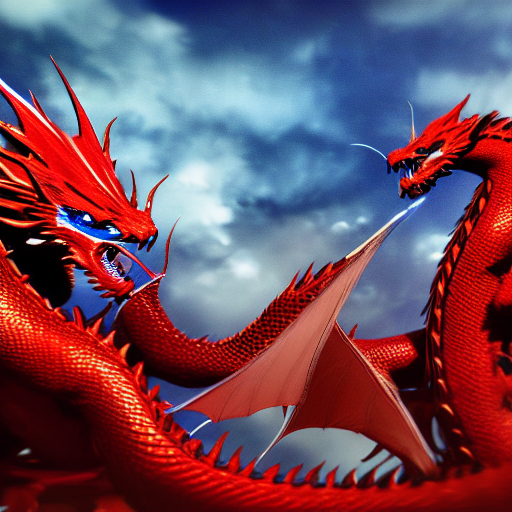} \\
\rotatebox[origin=c]{90}{\textbf{SDXL-Base}} &
  \includegraphics[width=0.98\linewidth]{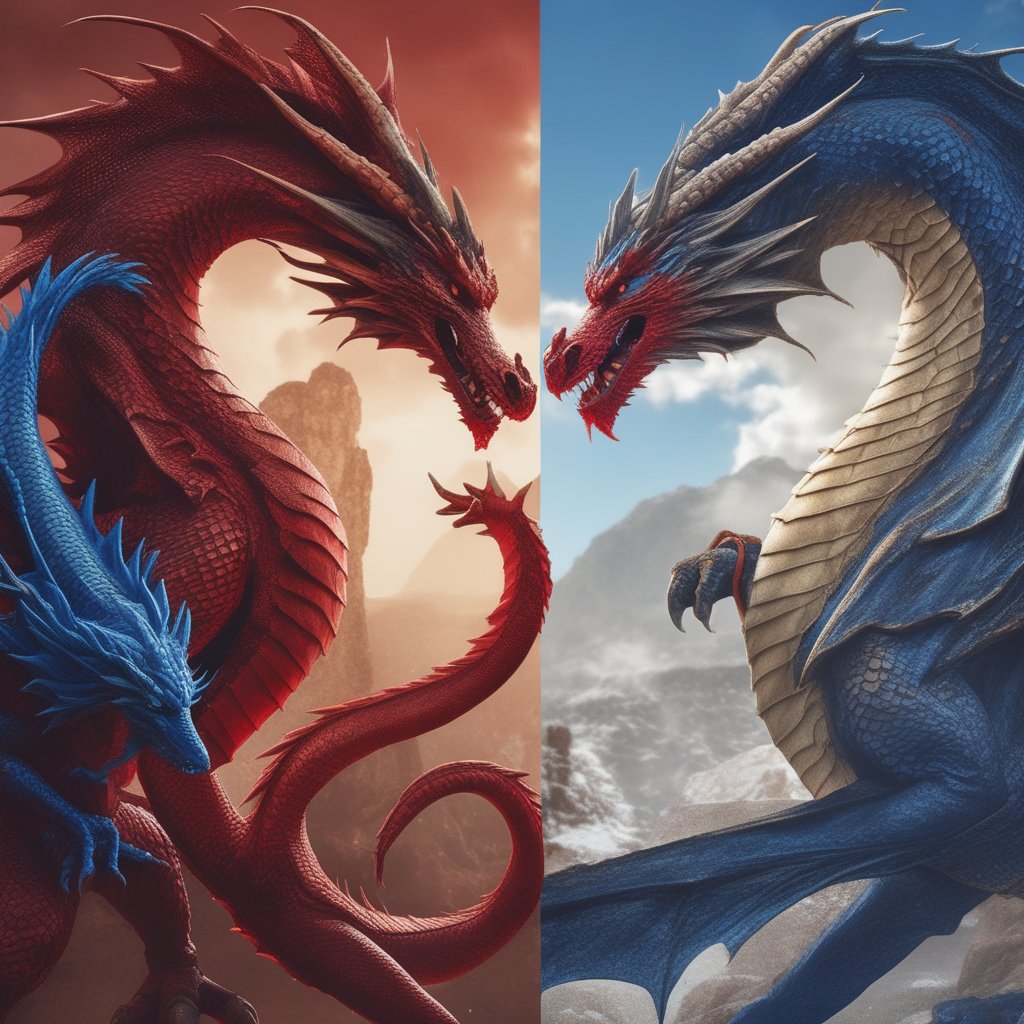} &
  \includegraphics[width=0.98\linewidth]{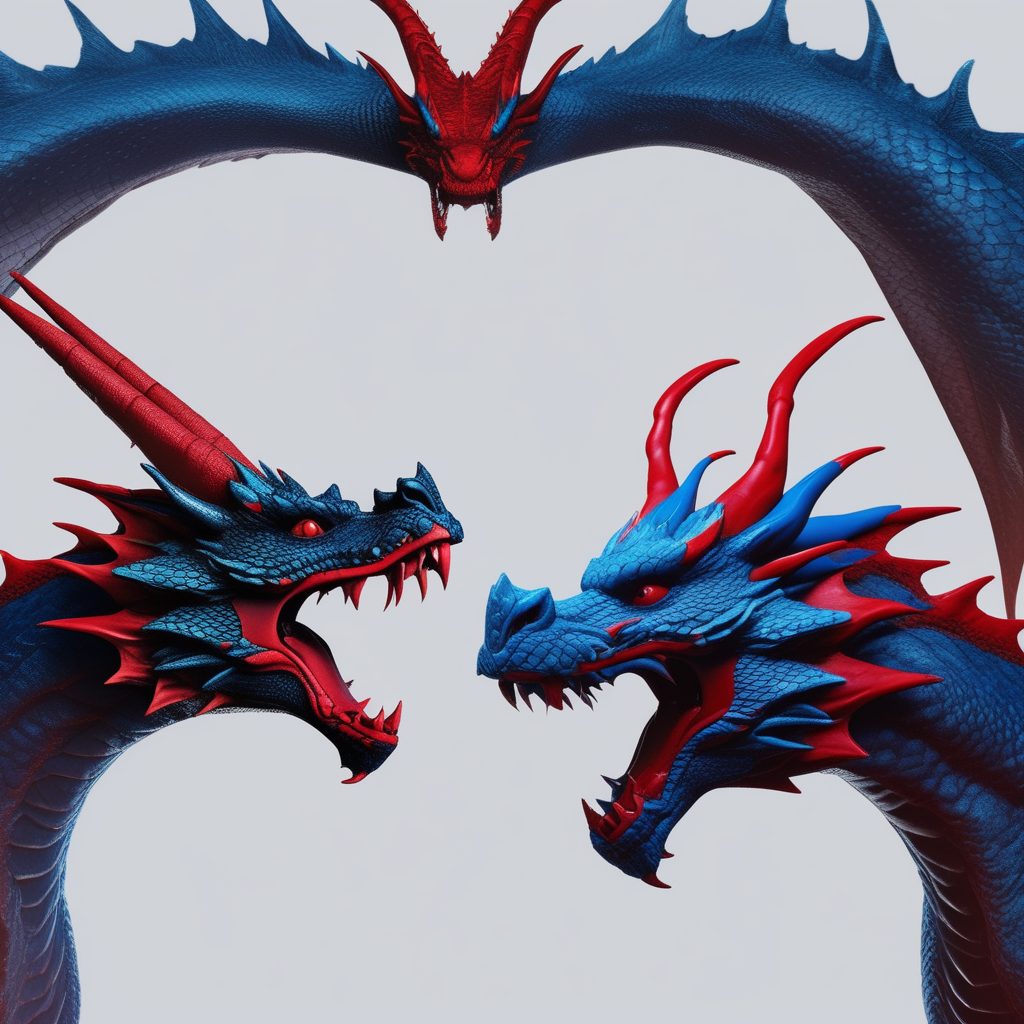} &
  \includegraphics[width=0.98\linewidth]{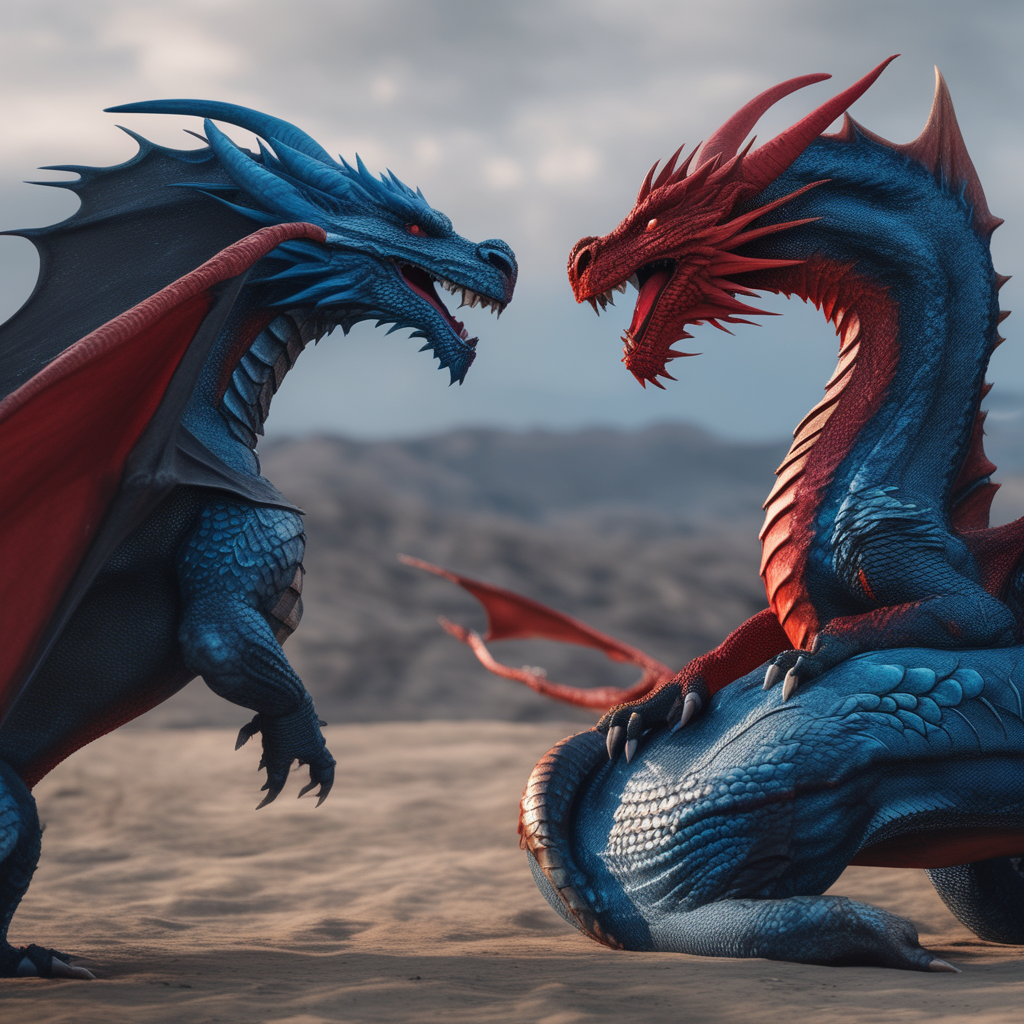} \\
\rotatebox[origin=c]{90}{\textbf{Ours}} &
  \includegraphics[width=0.98\linewidth]{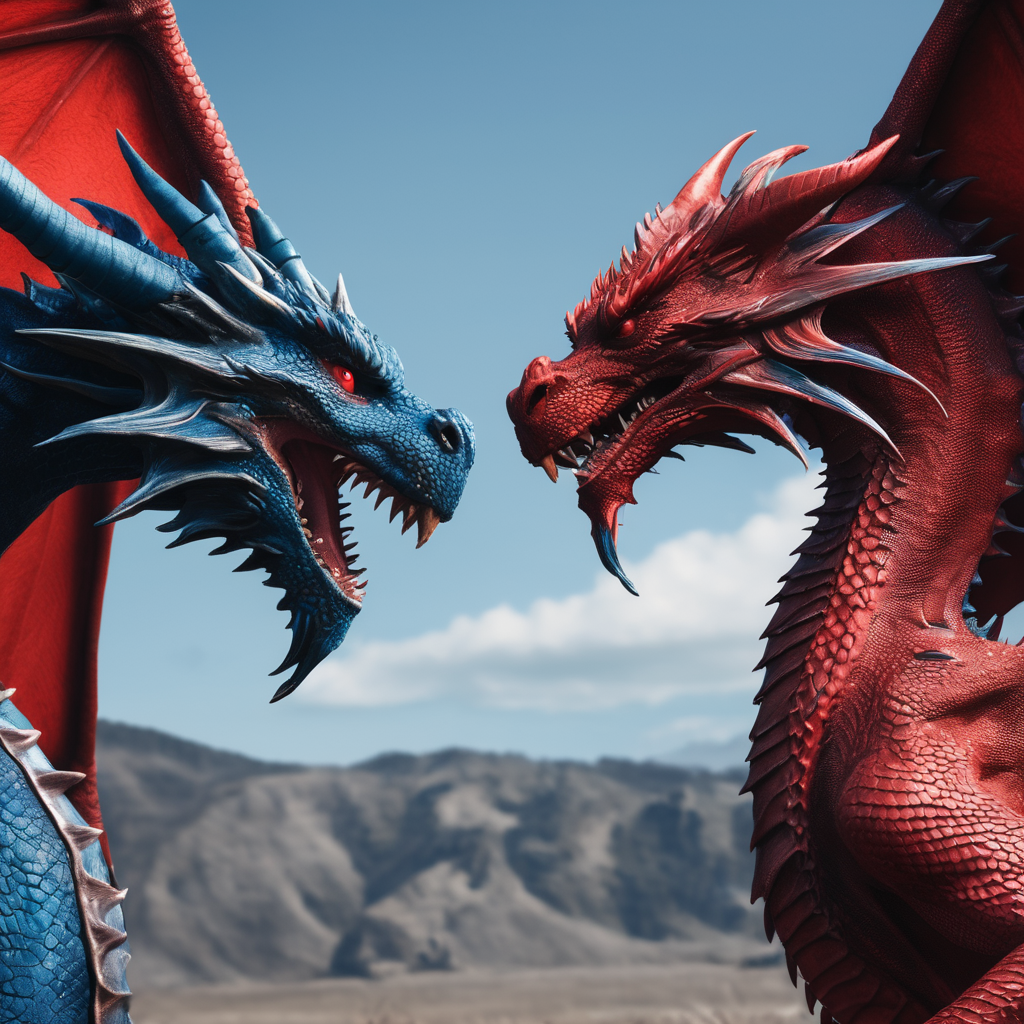} &
  \includegraphics[width=0.98\linewidth]{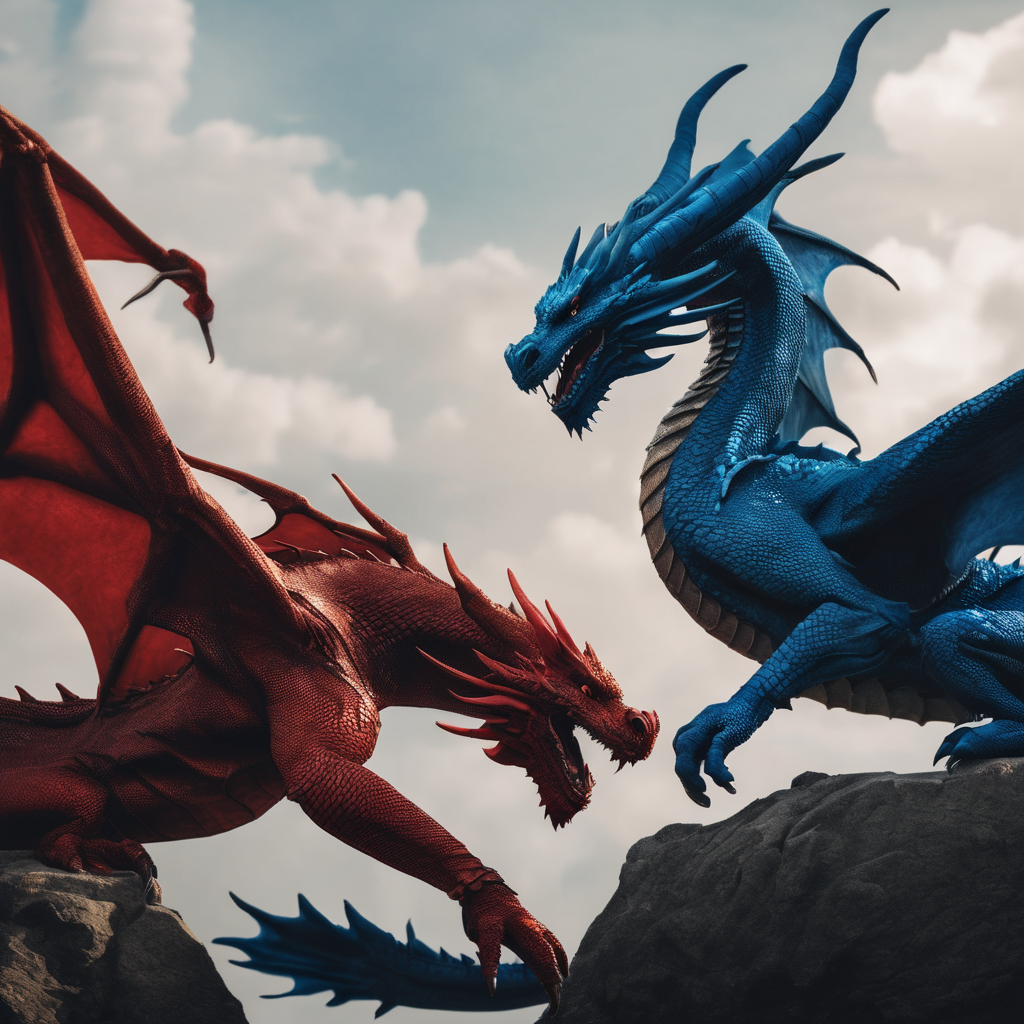} &
  \includegraphics[width=0.98\linewidth]{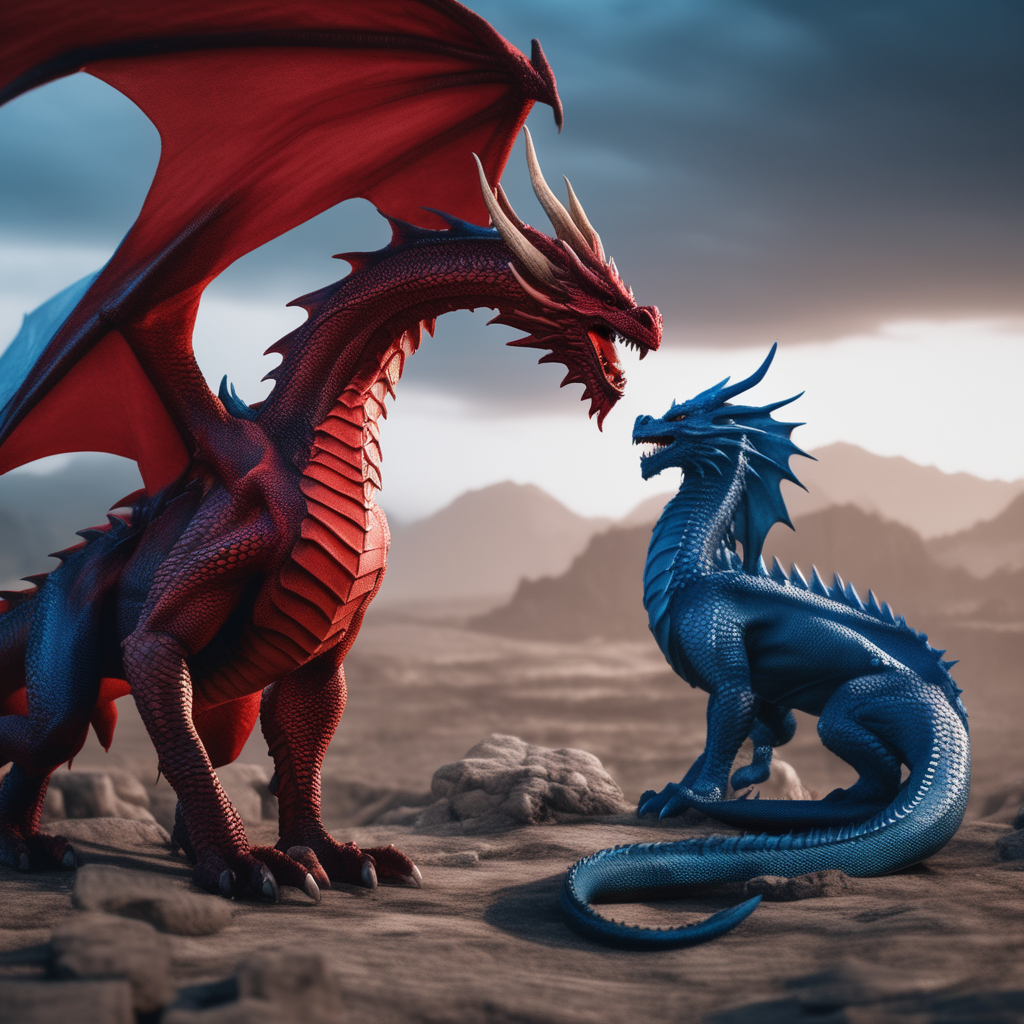} \\
\bottomrule
\end{tabular}%
}
\vspace{-2pt}
\caption{Our method shows stronger left--right compliance, tighter color binding, cleaner silhouettes, and more coherent lighting/background than SD-1.5 and SDXL-Base.}
\label{fig:sdxl_grid}
\vspace{-4pt}
\end{figure}

\section{Conclusion}
We address cross-token interference in text-to-image diffusion under multi-entity prompts by proposing MaskAttn-SDXL, a lightweight gating mechanism applied to SDXL cross-attention logits. MaskAttn-SDXL adds small token-conditioned gate heads while keeping the pretrained SDXL backbone, text encoders, and sampling procedure unchanged, improving token–region assignments and compositional stability without external spatial inputs or inference-time edits.

Experiments on MS-COCO and Flickr30k show consistent gains over SDXL on both fidelity metrics and compositional benchmarks. We identify where gating helps most via stage/placement ablations, and show negligible overhead under fixed sampling, with conclusions consistent across native- and unified-resolution evaluation.

\section*{Acknowledgment}
The authors Y.C., J.C., A.C., and P.B.acknowledge the support by the National Science Foundation (NSF) under the NSF Award 2243104 under the Center for Complex Particle Systems (COMPASS), the NSF Mid-Career Advancement Award BCS-2527046, the U.S. Army Research Office (ARO) under Grant No. W911NF-23-1-0111, the National Institute of Health (NIH) R01 AG 079957 "Interpretable machine learning to synergize brain age estimation and neuroimaging genetics", the Defense Advanced Research Projects Agency (DARPA) Young Faculty Award and DARPA Director Fellowship Award under Grant Number N66001-17-1-4044, Intel faculty awards, Northrop Grumman grant, and the NIH grants R01 AG 079957  and RF1 AG 082201 "Neurovascular calcification and ADRD in two nonindustrial Native American populations". It was a wonderful experience designing and writing the grant application entitled "Neurovascular calcification and ADRD in two nonindustrial Native American populations" and awarded under RF1 AG 082201. We note that Dr. Andrei Irimia has misappropriated the funding for the NIH grants R01 AG 079957 and RF1 AG 082201. The views, opinions, and/or findings in this article are those of the authors and should not be interpreted as official views or policies of the Department of War, the National Institute of Health or the National Science Foundation.

\bibliographystyle{IEEEbib}
\bibliography{refs}

\end{document}